\documentclass[10pt, a4paper]{article}
\usepackage{lrec}
\usepackage{graphicx}
\usepackage{tabularx}
\usepackage{soul}
\usepackage{booktabs}
\usepackage{multirow}
\usepackage{dblfloatfix}
\usepackage{epstopdf}
\usepackage[utf8]{inputenc}
\usepackage[hidelinks]{hyperref}
\usepackage{xstring}
\usepackage{caption,graphicx,newfloat}
\usepackage[dvipsnames]{xcolor}
\usepackage{titlesec}
\titleformat{\section}{\normalfont\large\bf\center}{\thesection.}{1em}{}
\titleformat{\subsection}{\normalfont\SmallTitleFont\bf\raggedright}{\thesubsection.}{1em}{}
\titleformat{\subsubsection}{\normalfont\normalsize\bf\raggedright}{\thesubsubsection.}{1em}{}
\renewcommand\thesection{\arabic{section}}
\renewcommand\thesubsection{\thesection.\arabic{subsection}}
\renewcommand\thesubsubsection{\thesubsection.\arabic{subsubsection}}

\title{Abstractive Text Summarization based on\\ Language Model Conditioning and Locality Modeling}

\name{\parbox{\textwidth}{\centering Dmitrii Aksenov, Juli\'{a}n Moreno-Schneider, Peter Bourgonje,\\ Robert Schwarzenberg, Leonhard Hennig, Georg Rehm}\vspace*{2mm}}
\address{DFKI GmbH, Alt-Moabit 91c, 10559 Berlin, Germany \\
\{firstname.lastname\}@dfki.de\\} 

\abstract{We explore to what extent knowledge about the pre-trained language model that is used is beneficial for the task of abstractive summarization. To this end, we experiment with conditioning the encoder and decoder of a Transformer-based neural model on the BERT language model. In addition, we propose a new method of BERT-windowing, which allows chunk-wise processing of texts longer than the BERT window size. We also explore how locality modeling, i.\,e., the explicit restriction of calculations to the local context, can affect the summarization ability of the Transformer. This is done by introducing 2-dimensional convolutional self-attention into the first layers of the encoder. The results of our models are compared to a baseline and the state-of-the-art models on the CNN/Daily Mail dataset. We additionally train our model on the SwissText dataset to demonstrate usability on German. Both models outperform the baseline in ROUGE scores on two datasets and show its superiority in a manual qualitative analysis. \\ \newline
\Keywords{Summarisation, Language Modeling, Information Extraction, Information Retrieval, BERT, Locality Modeling}}

\begin{document}

\maketitleabstract

\section{Introduction}
\label{sec:introduction}

Text summarization is an NLP task with many real-world applications. The ever-increasing amount of unstructured information in text form calls for methods to automatically extract the relevant information from documents and present it in condensed form. Within the field of summarization, different paradigms are recognised in two dimensions: extractive vs.~abstractive, and single-document vs.~multi-document. In extractive summarization, those sentences or words are extracted from a text which carry the most important information, directly presenting the result of this as the summary. Abstractive summarization methods paraphrase the text, and by changing the text aim to generate more flexible and consistent summaries. Furthermore, single-document summarization works on single documents, while multi-document summarization deals with multiple documents at once and produces a single summary. In this paper, we concentrate on single-document abstractive summarization. Most recent abstractive models utilize the neural network-based sequence-to-sequence approach. During training, such models calculate the conditional probability of a summary given the input sequence by maximizing the loss function (typically cross-entropy). Most approaches are based on the encoder-decoder framework where the encoder encodes the input sequence into a vector representation and the decoder produces a new summary given the draft summary (which is the part of the summary generated during previous iterations). The last layer of a decoder, the generator, maps hidden states to token probabilities. We use a state-of-the-art Transformer for sequence-to-sequence tasks which is built primarily on the attention mechanism \cite{Vaswani2017}.

We attempt to improve performance of abstractive text summarization by improving the language encoding capabilities of the model. Recent results have shown that the main contribution of the Transformer is its multi-layer architecture, allowing Self-Attention to be replaced with some other technique without a significant drop in performance \cite{domhan-2018-much,wu2018pay}. Following this strategy, we develop a model that introduces convolution into the vanilla Self-Attention, allowing to better encode the local dependencies between tokens. 
To overcome the data sparsity problem, we use a pre-trained language model for the encoding part of the encoder-decoder setup, which creates a contextualized representation of the input sequence. Specifically, we use BERT due to its bi-directional context conditioning, multilingualism and state-of-the-art scores on many other tasks \cite{devlin2018}. Furthermore, we propose a new method which allows applying BERT on longer texts. The main contributions of this paper are: (1) Designing two new abstractive text summarization models based on the ideas of conditioning on the pre-trained language model and application of convolutional self-attention at the bottom layers of the encoder. (2) Proposing a method of encoding the input sequence in windows which alleviates BERT's input limitations\footnote{BERT can process sequences with a maximum of 512 tokens.} and allows the processing of longer input texts. (3) Evaluating the performance of our models on the English and German language by conducting an ablation study on CNN/Dail Mail and SwissText datasets and comparing it with other state-of-the-art methods.

\section{Related Work}
\label{sec:related}

\subsection{Pre-trained Language Models}
\label{sec:langmodels}

Traditionally, non-contextualized embedding vectors were used for pre-training neural-based NLP models \cite{mikolov2013distributed,pennington2014glove}. Recently, pre-trained language models exploiting contextualized embeddings, such as ELMo, GPT-2, BERT and XLNet raised the bar in many NLP tasks \cite{Peters:2018,Radford2019LanguageMA,devlin2018,yang2019xlnet}. Recent attempts to use these models for text summarization demonstrated their suitability by achieving new state-of-the-art results \cite{zhang2019pretrainingbased,liu2019finetune,Liu_2019}.

\subsection{Neural Abstractive Text Summarization}
\label{sec:relatedneural}

The neural approach toward abstractive summarization was largely adopted by state-of-the-art models \cite{shi2018neural}. A significant contribution was the pointer Generator Network \cite{See2017}. It uses a special layer on top of the decoder network to be able to both generate tokens from the dictionary and extract them from the input text. It uses the coverage vector mechanism to pay less attention to tokens already covered by previous iterations. An example of earlier work adapting Reinforcement Learning (RL) is described by \newcite{paulus2017deep}. The pure RL model achieved high ROUGE-1 and ROUGE-L scores but produced unreadable summaries. Its combination with typical cross-entropy optimization achieved high scores eliminating the unreliability problem. \newcite{Liu2018}, to the best of our knowledge, were the first to use the Transformer model for summarization. It was only used in the decoder on top of the extraction model with various attention compression techniques to increase the size of the input sequence. \newcite{zhang2019pretrainingbased} incorporate BERT into the Transformer-based model. They use a two-stage procedure exploiting the mask learning strategy. Others attempt to improve their abstractive summarization models by incorporating an extractive model. For example, \newcite{li-etal-2018-guiding} use the Key information guide network to guide the summary generation process. In Bottom-up summarization \cite{gehrmann2018bottom} the extractive model is used to increase the precision of the Pointer Generator mechanism. Another strand of research adapts existing models to cope with long text. \newcite{Cohan2018ADA} present the Discourse-Aware Attention model which introduces hierarchy in the attention mechanism via calculating an additional attention vector over the sections of the input text. \newcite{s2019extractive} showed that the language model trained on the combination of the original text, extractive summaries generated by the model and the golden summary can achieve results comparable to standard encoder-decoder based summarization models.

\section{Approach}
\label{sec:approach}

Our text summarization model is based on the Transformer architecture. This architecture adopts the original model of \newcite{Vaswani2017}. On top of the decoder, we use a Pointer-Generator (Formula~\ref{copy}) to increase the extractive capabilities of the network (we later refer to this architecture as CopyTransformer). 
\begin{equation}
p(w) = p_{gen}  P_{copy}(w)  +  (1-p_{gen})  P_{softmax}(w) ,
\label{copy}
\end{equation}
where $P_{copy}(w)$ is the probability of copying a specific word $w$ from the source document, $P_{softmax}(w)$ is the probability of generation a word calculated by the abstractive summarization model and $p_{gen}$ is the probability of copying instead of generation.

\begin{figure}[t]
  \centering
  \includegraphics[width=.8\columnwidth]{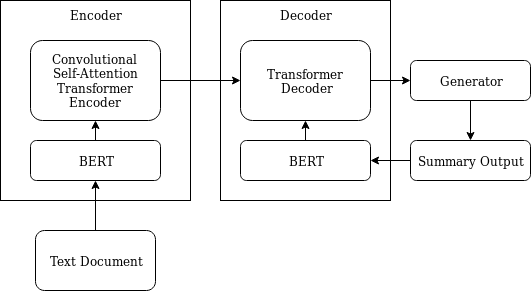}
  \caption{Model overview}\label{fig:summmodel}
\end{figure}

\subsection{Convolutional Self-Attention}
\label{sec:localatten}

The Transformer, like any other self-attention network, has a hierarchical multi-layer architecture. In many experiments it was shown that this architecture tends to learn lexical information in the first layers, sentence-level patterns in the middle and the semantics in the upper layers \cite{raganato-tiedemann-2018-analysis,Tenney_2019}. 
The disadvantage of this approach is that during the attention operation it considers all tokens as equally important, whereas syntactic information is mostly concentrated in certain local areas. This problem is usually specified as the problem of locality modeling. As syntactic information can help in identifying more important words or phrases, it could be beneficial to focus attention on these regions.

A successful approach to the locality modeling task are the so-called convolutions (local) self-attention networks \cite{localattention}. Essentially, the problem is dealt with by the application of a 1-dimensional convolution to the self-attention operation at the network's lower layers. This strengthens dependencies among neighboring elements and makes the model distance-aware when it searches for low-level patterns in a sequence. In other words, it restricts the attention scope to the window of neighboring elements. The 1D convolution applied to attention is illustrated in Formulas~\ref{eqn-1D_key}, \ref{eqn-1D_value} and~\ref{eq:conc}.

\begin{equation}
    \widehat{\mathbf{K}}^h = \{\mathbf{k}^h_{i-\frac{M}{2}}, \dots, \mathbf{k}^h_i, \dots, \mathbf{k}^h_{i+\frac{M}{2}} \} ,
    \label{eqn-1D_key}
\end{equation}
\begin{equation}
    \widehat{\mathbf{V}}^h = \{\mathbf{v}^h_{i-\frac{M}{2}}, \dots, \mathbf{v}^h_i, \dots, \mathbf{v}^h_{i+\frac{M}{2}} \}  ,
    \label{eqn-1D_value}
\end{equation}
\begin{equation}
    \mathbf{o}^h_i = \textsc{Att}(\mathbf{q}^h_i, \widehat{\mathbf{K}}^h) \widehat{\mathbf{V}}^h ,
    \label{eq:conc}
\end{equation}
where $\mathbf{q}^h_i$ is the query and $M+1$ ($M \le I$) is its attention region centered at the position $i$.
\\

The convolution can be extended to the 2-dimensional area by taking interactions between features learned by the different attention heads of the Transformer into account. In the original Transformer each head independently models a distinct set of linguistic properties and dependencies among tokens \cite{raganato-tiedemann-2018-analysis}. By applying 2-dimensional convolution, where the second dimension is the index of attention head, we explicitly allow each head to interact with learned features for their adjacent sub-spaces. The shortcoming of the original implementation is that the first and the last heads do not interact as they are assumed not to be adjacent. Thus, we assume that considering the heads' sub-spaces periodically, we can increase the model's effectiveness by applying circular convolution to the second dimension. In Section~\ref{sec:experiments}, we evaluate both the original version and our modification.

\begin{equation}
    \widetilde{\mathbf{K}}^h =  \bigcup [\widehat{\mathbf{K}}^{h-\frac{N}{2}}, \dots, \widehat{\mathbf{K}}^{h}, \dots,  \widehat{\mathbf{K}}^{h+\frac{N}{2}}] , 
\end{equation}
\begin{equation}
    \widetilde{\mathbf{V}}^h =  \bigcup [\widehat{\mathbf{V}}^{h-\frac{N}{2}}, \dots, \widehat{\mathbf{V}}^{h}, \dots, \widehat{\mathbf{V}}^{h+\frac{N}{2}}] , 
\end{equation}
\begin{equation}
    \mathbf{o}^h_i = \textsc{Att}(\mathbf{q}^h_i, \widetilde{\mathbf{K}}^h) \widetilde{\mathbf{V}}^h , 
\end{equation}
where $(M+1)$ ($N \le H$) is the window region over heads and $\bigcup$ stands for the union of keys $\widehat{\mathbf{K}}^h$ and values $\widehat{\mathbf{V}}^h$ from different subspaces.
\\

The convolutional self-attention has been shown to be very effective in Machine Translation and several other NLP tasks. However, to our knowledge, it was never applied to the text summarization problem. For the experiments reported on in this paper, we created our implementation of the local attention and the convolutional self-attention network (Transformer). It supports both 1D and 2D modes having the size of the kernels as system parameters. As in \newcite{localattention} we incorporate convolutional self-attention in the Transformer encoder by positioning it in the place of the self-attention in the lower layers. In Section~\ref{sec:experiments}, we show that the low-level modeling capabilities of our encoder provides a strong boost to the model's prediction accuracy in the text summarization task. 

\subsection{BERT-Conditioned Encoder}
\label{sec:knowledgetransferencoder}

The main task of the encoder is to remember all the semantic and syntactic information from the input text which should be used by the decoder to generate the output. Knowledge transfer from the language model should theoretically improve its ability to remember the important information due to the much larger corpus used in its pre-training phase compared to the corpus used in the text summarization training phase. We thus condition our encoder on the BERT language model.

For the encoder conditioning, we used the most straightforward strategy recommended for the BERT based model: placing the pre-trained language model in the encoder as an embeddings layer. This should make the embeddings of the system context-dependent. We decided not to fine-tune the encoder on BERT for the sake of memory and time economy. Instead, we follow the general recommendations by concatenating the hidden states of the last four layers of BERT into a 3072-dimensional embedding vector \cite{devlin2018}. We use two variations of the BERT-based encoder. The first model uses only BERT to encode the input sequence and the second model feeds BERT's generated embeddings into the vanilla Transformer encoder. 

\subsection{BERT-Windowing}
\label{sec:windowing}

\begin{figure}[ht]
  \centering
  \includegraphics[width=.7\columnwidth]{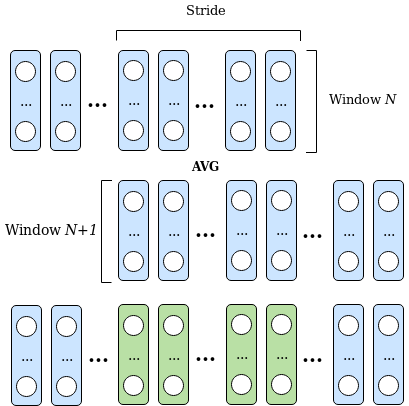}
  \caption{Integration of BERT-generated contextual representations from two windows}\label{fig:windowing}
\end{figure}

One of the key features of our approach is its ability to overcome the length limitations of BERT, allowing it to deal with longer documents. BERT's maximum supported sequence length is 512 tokens\footnote{These are not tokens in the traditional sense, but so-called WordPiece tokens, see \newcite{devlin2018}.}, which is smaller than the average size of texts used in most summarization datasets. Our method relies on the well-known method of windowing which to our knowledge was never used before neither in the BERT-based models nor in abstractive text summarization research (Figure~\ref{fig:windowing}). We apply BERT to the windows of texts with strides and generate $N$ matrices, every matrix embedding one window. Then we combine them by doing the reverse operation. The vectors at the overlapping positions are averaged (by summing and dividing by the number of overlapping vectors). As a result, we have the matrix of embeddings with the shape of the hidden size times the length of the text. The drawback of this approach is that we reduce the size of the context as each resulted vector is calculated based on maximum twice the window size number of tokens. Besides, the split of the text to equal size windows will aggravate the consistency of the input as some sentences will be split in an arbitrary manner between two adjacent windows. Despite this drawback, we assume that this procedure will nevertheless improve the accuracy of the encoder trained on the non-truncated texts. We set the window size to the maximum size of 512 tokens and the stride to 256. We consider this stride size optimal due to a trade-off between the average context size and computational requirements of the model (number of windows). By this trade we ensure every token to have a 768 tokens-context except for the 256 initial and final tokens, that only have 512 tokens-context.

\begin{figure*}[t]
  \centering
  \includegraphics[scale=0.45]{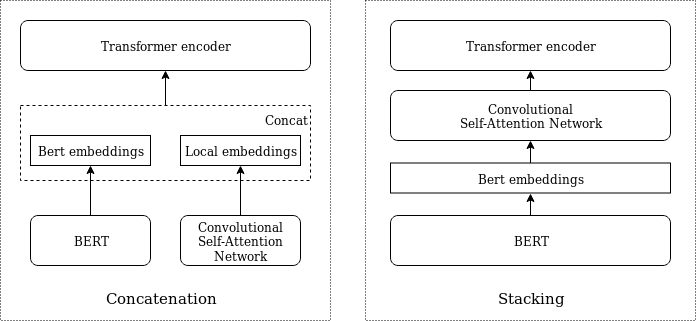}\\
  \caption{Two different approaches for the integration of the BERT-conditioning with Convolutional Self-Attention}
  \label{fig:integration}
\end{figure*}

\subsection{BERT-Conditioned Decoder\label{sec:knowledgetransferdecoder}}

In the decoder, pre-training was applied in a similar way. The main difference is that instead of the final output of BERT we use only its word embedding matrix (without positions). The reason behind this is that in the decoder the generated probability distribution is conditioned on the incomplete text (previous summary draft output) while BERT implicitly assumes consistent and completed input \cite{zhang2019pretrainingbased}. As context-independent embeddings are not enough to represent the minimum set of features to make a meaningful prediction, the custom Transformer decoder is always stacked on top of BERT.

Our whole BERT-based model is similar to One-Stage BERT \cite{zhang2019pretrainingbased} and BertSumAbs \cite{Liu_2019} but differs in the usage of the four last hidden states of BERT to create contextualized representation, in presence of Pointer Generator and capabilities to process long texts. In Figure~\ref{fig:summmodel} we show the schema of the basic model with the BERT-conditioned convolutional self-attention encoder and BERT-conditioned decoder.

\subsection{Integration of BERT and Convolutional Self-Attention}
\label{sec:encoderintegratin}

We evaluated two different ways of integrating the BERT-conditioning with the convolutional self-attention of the model's encoder (Figure~\ref{fig:integration}).

\paragraph{Stacking} This approach comprises feeding the BERT-generated embeddings to the convolutional self-attention Transformer encoder. A potential problem with this approach is that convolutional self-attention is assumed to be beneficial when applied in the lower layers as its locality modeling feature should help in modeling of local dependencies (e.\,g., syntax). At the same time, BERT is a hierarchical model where the last layers target high-level patterns in the sequences (e.\,g., semantics). We assume that the application of the network detecting the low-level patterns on BERT's output can undermine its generalization abilities. 

\paragraph{Concatenation} Because of the considerations raised above, we also develop a second approach which we call Concatenation. We split the convolutional self-attention Transformer encoder into two networks where the first one uses only convolutional self-attention and the second original self-attention (identical to the Transformer encoder). Then we feed the original sequences into BERT and into the convolutional self-attention network in parallel. The resulting embedding vectors are concatenated and fed into the Transformer encoder. In this way, we model the locality at the lower layers of the encoder at the cost of a smaller depth of the network (assuming the same number of layers).

\begin{table}[t]
\centering\small
\begin{tabular}{@{}lccc@{}} \toprule
Method        & ROUGE-1 & ROUGE-2 & ROUGE-L \\ \midrule
CopyTransformer & 31.95	& 14.49	& 30.02 \\ \midrule
+ 1D conv. & 32.62 & 14.99	& 30.74\\
+ 2D conv. & \textbf{32.72} & \textbf{15.12} & \textbf{30.85}\\
+ 2D Circular conv. & 32.68 & 15.01 & 30.76 \\ \bottomrule
\end{tabular}
\caption{Ablation study of model with Convolutional Self-Attention on the CNN/Daily Mail dataset (kernel sizes are 11 and 3)}
\label{tab:conv}
\end{table}

\begin{figure}[b]
  \centering
  \includegraphics[width=\columnwidth]{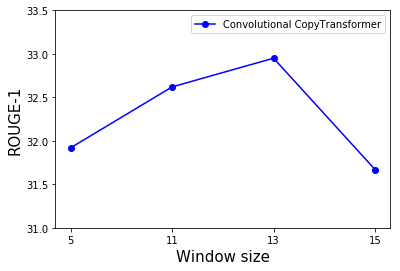}
  \caption{Effect of the window size on ROUGE-1}
  \label{fig:window}
\end{figure}

\section{Datasets}
\label{datests}

We aim to develop a system that works in a language-independent way. It assumes that either the upstream components are available in the respective language, or they are themselves language-independent, such as the multi-lingual version of BERT. Since most summarization datasets are in English however, we use English for the evaluation and additionally include German to check if of our model can be applied to another language.

\begin{table*}[t]
\centering\small
\begin{tabular}{@{}lrrr@{}} \toprule
Model 				            & ROUGE-1 & ROUGE-2 & ROUGE-L \\ \midrule
Transformer & 24.82 & 6.27 & 22.99 \\
CopyTransformer & 31.95 & 14.49 & 30.02\\
Bert Encoder + Transformer Decoder & 31.3 & 13.37	& 29.46\\
Bert-transformer Encoder + Transformer Decoder & 32.5 & 14.68	& 30.68\\
Bert-transformer Encoder + Bert-transformer Decoder & \textbf{33.23} & \textbf{14.99}	& \textbf{31.26}\\ \midrule
Transformer (full text) & 23.18 & 5.15 & 21.48\\
Bert-transformer Encoder + Transformer Decoder (full text) & \textbf{31.51} & \textbf{14.1} & \textbf{29.77}\\ \bottomrule
\end{tabular}
\caption{Ablation study of the BERT-based model on truncated and original CNN/Daily Mail dataset}
\label{tab:bert}
\end{table*}

\begin{table*}[t]
\centering\small
\begin{tabular}{@{}lrrr@{}} \toprule
Model 				            & ROUGE-1 & ROUGE-2 & ROUGE-L \\ \midrule
Transformer & 36.40 & 20.69 & 34.14\\
CopyTransformer & 39.44 & 25.11 & 37.16\\
Bert-transformer Encoder + Transformer Decoder & \textbf{44.01} & \textbf{29.60} & \textbf{41.65}\\
Bert-transformer Encoder + Bert-transformer Decoder & 43.22 & 29.01 & 40.84\\ \midrule
Transformer (full text) & 34.76 & 18.65 & 32.61\\
Bert-transformer Encoder + Transformer Decoder (full text) & \textbf{45} & \textbf{30.49} & \textbf{42.64}\\ \bottomrule
\end{tabular}
\caption{Ablation study of the BERT-based model on the truncated and original SwissText dataset}
\label{tab:bertswiss}
\end{table*}

\subsection{CNN/Daily Mail}
\label{sec:cnndm}

Our experiments are mainly conducted on the CNN/Daily Mail dataset \cite{Hermann2015,Nallapati2016}. It contains a collection of news articles paired with multi-sentence summaries published on the CNN and Daily Mail websites. This dataset is the de facto standard for training summarization models. We use the non-anonymized data as was used for training of the most recent state-of-the-art models (e.\,g., \newcite{See2017}). The raw dataset consists of separate text files each representing a single article or a summary. We use the data in its preprocessed version as provided by \newcite{gehrmann2018bottom}. It has 287,226 training pairs, 13,368 validation pairs and 11,490 test pairs. 

To align the data with the vocabulary of BERT we tokenized it using the BPE-based WordPiece tokenizer \cite{devlin2018}. As all samples in BERT's training data are prepended with the special token "[CLS]", we follow this and add it to every source text in our dataset. In the resulting dataset, the average lengths of an article and a summary are 895 and 63 tokens, respectively. In most of our experiments, we use the clipped version of the training and validation datasets with each article truncated to 512 tokens. In the experiments on BERT windowing, we use the full-text version. 

\subsection{SwissText Dataset}
\label{sec:gerwiki}

To evaluate the efficiency of the model in a multi-lingual, multi-domain environment we conduct a series of experiments on the German SwissText dataset. This dataset was created for the 1st German Text Summarization Challenge at the 4th Swiss Text Analytics Conference -- SwissText 2019 \cite{Swisstext}. It was designed to explore different ideas and solutions regarding abstractive summarization of German texts. To the best of our knowledge, it is the first long document summarization dataset in the German language that is publicly available. The data was extracted from the German Wikipedia and represents mostly biographical articles and definitions of various concepts. 

The dataset was tokenized by the multilingual WordPiece tokenizer \cite{devlin2018} and preprocessed in the same way as the CNN/Daily Mail dataset. It was split into the training, validation and testing sets containing 90,000, 5,000 and 5,000 samples, respectively. The average length of a source sequence is 918 tokens, which makes this dataset suitable for our experiments on windowing.

\begin{table*}[b]
\centering\small
\begin{tabular}{@{}llrrr@{}} \toprule
Method of Integration & Model 				            & ROUGE-1 & ROUGE-2 & ROUGE-L \\  \midrule
\multirow{2}{*}{Stacking} & BERT+CopyTransformer & 35.28 & \textbf{17.12} & 33.31\\
& BERT+Convolutional CopyTransformer & \textbf{35.4} & 16.82 & \textbf{33.31}\\ \midrule
\multirow{2}{*}{Concatenation} & BERT+CopyTransformer & 34.82 & 16.46 & 32.79 \\
& BERT+Convolutional CopyTransformer & \textbf{35.26} & \textbf{16.79} & \textbf{33.22}\\ \bottomrule
\end{tabular}
\caption{Different strategies for integrating language models with convolutional Self-Attention (CNN/Daily Mail dataset)}
\label{tab:integration}
\end{table*}

\section{Experiments}
\label{sec:experiments}

Our system is built on the OpenNMT library. For training, we use cross-entropy loss and the Adam optimizer with the Noam decay method \cite{kingma2014adam}. Regularization is made via dropout and label smoothing. For evaluation, we calculate the F1-scores for ROUGE using the files2rouge library. The ROUGE evaluation is made on the sequences of WordPiece tokens.

\subsection{Locality Modeling}
\label{sec:localmodel}

To evaluate the effect of convolution on self-attention we introduce it in the first layer of the encoder. We use the same kernel sizes as in \newcite{localattention}. In these experiments, to accelerate the training process, we use a small model with a hidden size of 256, four self-attention heads and three layers in the encoder and decoder. All models are trained for 90,000 training steps with the Coverage Penalty. As a baseline, we use our implementation of CopyTransformer. In contrast to \newcite{See2017}, we do not re-use the attention layer for the decoder but train a new Pointer-Generator layer from scratch. 

The results are presented in Table~\ref{tab:conv}. We see that both convolutions over tokens and over attention heads improve the ROUGE scores. Standard convolution outperformed circular convolution on ROUGE-1, ROUGE-2 and ROUGE-L by 0.06, 0.13 and 0.09 percent, respectively.

We also investigated the effect of the window size of the 1-dimensional convolution on ROUGE scores (Figure~\ref{fig:window}). In contrast to findings in Machine Translation, we found that size 13 returns the best result for the summarization task.

\subsection{BERT Conditioning}
\label{sec:bertcond}

To find the optimal architecture of the BERT-based abstractive summarizer we conducted an ablation study (Table~\ref{tab:bert}). All hyperparameters were set equal to the ones in experiments in convolutional self-attention. On CNN/Daily Main dataset we test three different models: BERT encoder+Transformer Decoder, BERT-Transformer encoder+Transformer decoder and BERT-Transformer encoder+BERT-Transformer decoder. The version of BERT used in the experiments is BERT-Base. As the baseline, we use the Transformer without Pointer Generator. From the results, we observe that BERT improves the efficiency of the model when it is used in both encoder and decoder. Besides, BERT in the encoder is more effective when it is used to produce embeddings to be used by the standard Transformer encoder than when it is used solely as an encoder. Even without a Pointer Generator, our model outperformed the CopyTransformer baseline by 1.28, 0.5 and 1.24 on ROUGE-1, ROUGE-2 and ROUGE-L.

To evaluate our BERT-windowing method we conducted the experiments on the full text. Our approach outperforms the baseline, which proves that the method can be successfully applied to texts longer than 512 tokens. The final performance of this model is still lower than that of the model trained on the truncated text, but as the same pattern can be observed for the baselines we assumed this relates to the specifics of the dataset that is prone to having important information in the first sentence of a text.

On SwissText data we use the multilingual version of BERT-Base. We evaluated two models with Bert-transformer encoder and Transformer and BERT-Transformer decoders (Table~\ref{tab:bertswiss}). The introduction of BERT into the transformer increased the ROUGE-1, ROUGE-2 and ROUGE-L scores by 7.21, 8.91 and 7.51 percent, respectively. At the same time, the usage of BERT in the decoder decreased the overall score. We assume that the reason behind this is that in multilingual BERT, due to its language-independence, the embedding matrix outputs less precise contextualized representations which undermines their benefits for the summarization task. 

On the non-truncated texts, usage of the Bert-transformer encoder increased the ROUGE scores by 10.23, 11.84 and 10.03 percent. Furthermore, it gives us higher scores compared to the same model on truncated texts. 
This demonstrates the usability of BERT-windowing for this particular dataset. We assume that the difference in performance on the CNN/Daily Mail datasets reflects the difference in distribution of the useful information within the text. Particularly, that in the SwissText dataset, it is spread more uniformly than in the CNN/Daily Mail dataset. We conducted a small experiment comparing the average ROUGE score between a golden summary and the head and the tail of a document (taking the first or last \textit{n} sentences, where \textit{n} correlates to the length of the gold summary) on both datasets. The difference between taking the head and a tail on the SwissText dataset (ROUGE-L of 34.79 vs. 20.15, respectively) was much smaller than on CNN / Daily Mail (ROUGE-L of 16.95 vs.~12.27, respectively) which confirms our hypothesis.

\subsection{Integration Strategies}
\label{sec:integration}

To evaluate the integration strategies, we trained two models with the respective BERT-based baselines. Both models have in their encoder two Transformer layers and one Convolutional Transformer layer placed on top of BERT or in parallel, respectively (Table~\ref{tab:integration}).

The method of stacking does not provide any significant improvement. With the introduction of convolutional self-attention only ROUGE-1 increased by 0.12 percent, while ROUGE-2 dropped by 0.3 and ROUGE-L remained the same. Considering that in many domains ROUGE-2 maximally correlates with human assessment (see Section~\ref{sec:evaluation}), we dismiss this method. The concatenation strategy convolution is shown to be much more efficient, increasing ROUGE scores by 0.44,0.33 and 0.43 percent. This confirms our hypothesis that locality modeling is the most efficient when applied at the bottom on the non-contextualized word representations. Unfortunately, this model failed to outperform the stacking baseline. We conclude that the concatenating architecture undermines the performance of the Transformer model, and the convolutional self-attention is not beneficial when used together with pre-trained language models. Hence, we decided to train our two final models separately.

\begin{table*}[t]
\centering\small
\begin{tabular}{@{}llll@{}} \toprule
Method 				            & ROUGE-1 & ROUGE-2 & ROUGE-L \\ \midrule
BiLSTM + Pointer-Generator + Coverage \cite{See2017} & 39.53 & 17.28 & 36.38 \\
ML + Intra-Attention \cite{paulus2017deep}   & 38.30 & 14.81 & 35.49 \\ 
CopyTransformer \cite{gehrmann2018bottom} & 39.25 & 17.54 & 36.45 \\  
Bottom-Up Summarization \cite{gehrmann2018bottom} & 41.22 & 18.68 & 38.34 \\
One-Stage BERT  \cite{zhang2019pretrainingbased} & 39.50 & 17.87 & 36.65 \\
Two-Stage BERT  \cite{zhang2019pretrainingbased} & 41.38 & 19.34 & 38.37\\
ML + Intra-Attention + RL \cite{paulus2017deep}      & 39.87 & 15.82 & 36.90 \\
Key information guide network \cite{li-etal-2018-guiding}  & 38.95 & 17.12 & 35.68 \\
Sentence Rewriting \cite{chen-bansal-2018-fast}  & 40.88 & 17.80 & 38.54 \\
BertSumAbs \cite{Liu_2019}  & \textbf{41.72} & \textbf{19.39} & \textbf{38.76}\\ \midrule
CopyTransformer (our implementation)  & 38.73 & 17.28 & 35.85\\  
Convolutional CopyTransformer & 38.98 & 17.69 & 35.97\\ 
BERT+CopyTransformer (enc., dec.) & \textbf{40} & \textbf{18.42} & \textbf{37.15}\\ \bottomrule
\end{tabular}
\caption{ROUGE scores for various models on the CNN/Daily Mail test set. The first section shows different state-of-the-art models, the second section presents our models and baseline.}
\label{tab:abs}
\end{table*}

\begin{table*}[t]
\centering\small
\begin{tabular}{@{}llll@{}} \toprule
Method 				             & ROUGE-1 & ROUGE-2 & ROUGE-L \\ \midrule
CopyTransformer (our implementation)   & 39.5 & 22.36 & 36.97\\ 
Convolutional CopyTransformer & 40.54 & 23.62 & 38.06\\ 
BERT+CopyTransformer (enc.) & \textbf{42.61} & \textbf{25.25} & \textbf{39.85}\\  \bottomrule
\end{tabular}
\caption{ROUGE scores for our models on the SwissText test set}
\label{tab:absswiss}
\end{table*}

\subsection{Model Comparison} 
\label{sec:finalmodel}

For the final comparison of our model to other state-of-the-art methods we conducted experiments on the CNN/Daily Mail dataset. We set the hidden state to 512, the number of Transformer layers in the encoder and layers to six and the number of self-attention heads to eight. Hence, our baseline is smaller than the original CopyTransformer \cite{gehrmann2018bottom}, which may be the reason why it performs slightly worse (Table~\ref{tab:abs}). BERT-conditioning was used in both the encoder and decoder. The sizes of convolution kernels are set to 13 and three. The networks were trained for 200,000 training steps on a single NVIDIA GeForce GTX 1080 Ti. The generation of the summary was made via the Beam search algorithm with the Beam size set to four. Finally, the generated summaries were detokenized back to the sequences of words separated by spaces. 

For the BERT-based model, we set the minimum length of a generated summary to 55 as we found that without such restriction the model was prone to generate shorter sequences than in the test dataset. The model outperformed the baseline by 1.27 on ROUGE-1, 1.14 on ROUGE-2 and 1.3 on ROUGE-L. This is better than the scores of One-Stage BERT but still worse than the two-stage and BertSumAbs models.

For the convolutional CopyTransformer we use convolutional self-attention in the first three layers of the encoder. It increased ROUGE-1, ROUGE-2 and ROUGE-L by 0.25, 0.41 and 0.12.

Furthermore, we present the first publicly available benchmark for the SwissData dataset (Table~\ref{tab:absswiss}).\footnote{For comparability with our other model we include results for the bigger BERT+CopyTransformer model. At the same time, we found that the smaller model without the copy mechanism achieved higher scores with 45.12 ROUGE-1, 28.38 ROUGE-2 and 42.99 ROUGE-L. This needs to be explored in future work.} All parameters are equal to the CNN/Daily Mail baseline. BERT-conditioning was used only in the encoder. The networks were trained on the truncated texts in 90,000 training steps. From the results we see that the convolutional CopyTransformer showed much more efficiency than on CNN/Daily Mail dataset, outperforming the baseline by 1.04 percent on ROUGE-1, 1.26 on ROUGE-2 and 1.09 on ROUGE-L. The BERT-based model achieved the highest scores.

\section{Qualitative Analysis}

\begin{figure*}[t]
\small
\noindent\fbox{%
    \parbox{484pt}{%
         \textbf{Gold summary:} researchers are developing a computer that can write weather forecasts .  it takes meteorological data and writes a report designed to mimic a human .  this process is known as ` natural language generation ' - lrb - nlg - rrb - .  a prototype system will be tested on the bbc website later this year .
    }%
}

\noindent\fbox{%
    \parbox{484pt}{%
        \textbf{Transformer:} researchers from london and edinburgh have developed a computer that can \colorbox{GreenYellow}{collateological} information . these computer - generated weather updates are being tested by scientists at heriot - watt university and university college london . if the project is successful , a prototype system will be tested by generating local weather reports on the bbc ' s website . currently , the bbc website features 10 reports written by meteorologists .
    }%
}
\noindent\fbox{%
    \parbox{484pt}{%
        \textbf{Convolutional Transformer:} researchers from london and edinburgh have developed a computer that can collate \colorbox{GreenYellow}{meterological} information and then produce forecasts as if they were written by a human . it uses a process known as ` natural language generation ' - lrb - nlg - rrb - . these computer - generated weather updates are being tested by scientists at heriot - watt university and university college london . if the project is successful , a prototype system will be tested by generating local weather reports on the bbc ' s website .
    }%
}
\noindent\fbox{%
    \parbox{484pt}{%
        \textbf{BERT-Transformer:} researchers from london and edinburgh have developed a computer that can collate meteorological information and produce forecasts as if they were written by a human . \colorbox{SkyBlue}{using met office data , it uses a process} known as ` natural language generation ' - lrb - nlg - rrb - . if the project is successful , a prototype system will be tested by generating local weather reports on the bbc ' s website .
    }%
}
    \caption{Comparison of the output of models on an example form CNN/Daily Mail testset. Surface realisation mistakes are highlighted in green and a typical abstractive feature, illustrating re-arranging of the sentence is highlighted in blue.\label{box:texts}}
\end{figure*}
\label{sec:testsamples}

As ROUGE evaluation is not always a valid method for quality assessment we perceive the need for an additional, manual evaluation. The best solution would be to conduct a fine-grained study of the models' outputs by manually ranking them in terms of semantic coherence, grammaticality, etc. However, due to the time-consuming nature of such an evaluation, we reverted to a qualitative analysis comparing several summaries generated by different models. Figure~\ref{box:texts} includes the reference summary and those generated by the different models. Comparing the first sentence we see that the vanilla Transformer model performed worse by copying only part of the original sentence omitting some characters in the word ``meteorological''. The model with convolution has copied the whole sentence but still made a spelling error. Finally, only the BERT-based model succeeded to generate the right token, ``meteorological''. Also, we see that while the BERT-based model's summary conveys the same meaning as the gold summary, the convolutional Transformer generates one and Transformer two sentences that are not present in the gold summary. Overall, on the given example all models provided a summary of extractive nature and only the BERT-based model shows some level of abstractiveness merging parts of the two sentences into the single one (in the second summary's sentence). This is far from the gold summary where every sentence in some way paraphrases the original text. Hence, given this particular example, our models demonstrate some explicit improvements. Still, abstractive summarization remains challenging. The paraphrasing capabilities of all state-of-the-art systems are low and the models are not guaranteed to produce summaries which follow the initial order of the sequence of events.

\section{Discussion: Summarization Evaluation}
\label{sec:evaluation}

ROUGE \cite{lin-2004-rouge} is the most widely adopted metric used for evaluating automatic text summarization approaches. The evaluation is made though comparison of a set of system-generated candidate summaries with a gold standard summary. The availability of the corresponding software and its performance contributed to its popularity \cite{cohan2016revisiting}. Despite its adoption in many studies, the metric faced some key criticisms.

The main criticism of ROUGE is that it does not take into account the meaning expressed in the sequences. The metric was developed based on the assumption that a high quality generated candidate summary should share many words with a single human-made gold standard summary. This assumption may be very relevant to extractive, but not to abstractive summarization, where different terminology and paraphrasing can be used to express the same meaning \cite{cohan2016revisiting}. This results in the metric assigning low scores to any summary not matching the gold standard on the surface level. This also allows cheating the metric by generating ungrammatical and nonsensical summaries having very high ROUGE scores. \newcite{SJOBERGH20071500} show how this can be achieved by choosing the most frequent bigrams from the input document.

ROUGE adoption relies on its correlation with human assessment. In the first research on the DUC and TDT-3 datasets containing news articles, ROUGE indeed showed a high correlation with the human judgments \cite{lin-2004-rouge,dorr-etal-2005-methodology}. However, more recent research questions the suitability of ROUGE for various settings. \newcite{conroy-dang-2008-mind} show that on DUC data the linguistic and responsiveness scores of some systems do not correspond to the high ROUGE scores. \newcite{cohan2016revisiting} demonstrate that for summarization of scientific texts, ROUGE-1 and ROUGE-L have very low correlations with the gold summaries. ROUGE-N correlates better but is still far from the ideal case. This follows the result of \newcite{murray}, showing that the unigram match between the candidate summary and gold summary is not an accurate metric to assess quality.

Another problem is that the credibility of ROUGE was demonstrated for the systems which operated in the low-scoring range. \newcite{peyrard-2019-studying} show that different summarization evaluation metrics correlate differently with human judgements for the higher-scoring range in which state-of-the-art systems now operate. Furthermore, improvements measured with one metric do not necessarily lead to improvements when using others.

This concern led to the development of new evaluation metrics. \newcite{peyrard-2019-simple} define metrics for important concepts with regard to summariazion: Redundancy, Relevance, and Informativeness in line with Shannon’s entropy. From these definitions they formulate a metric of Importance which better correlates to human judgments. \newcite{clark-etal-2019-sentence} propose the metric of Sentence Mover’s Similarity which operates on the semantic level and also better correlates with human evaluation. A summarization model trained via Reinforcement Learning with this metric as reward achieved higher scores in both human and ROUGE-based evaluation.

Despite these drawbacks, the broad adoption of ROUGE makes it the only way to compare the efficiency of our model with other state-of-the-art models. The evaluation of our system on the SwissData dataset confirms that its efficiency (in terms of ROUGE) is not restricted to CNN/Daily Mail data only.

\section{Conclusion}
\label{sec:conclusions}

We present a new abstractive text summarization model which incorporates convolutional self-attention in BERT. We compare the performance of our system to a baseline and to competing systems on the CNN/Daily Mail data set for English and report an improvement over state-of-the-art results using ROUGE scores. To establish suitability of our model to languages other than English and domains other than that of the CNN/Daily Mail data set, we apply our model to the German SwissText data set and present scores on this setup. A key contribution of our model is the ability to deal with texts longer than BERT's window size which is limited to 512 WordPiece tokens. We present a cascading approach and evaluate this on texts longer than this window size, and demonstrate its performance when dealing with longer input texts. 

The source code of our system is publicly available.\footnote{\url{https://github.com/axenov/BERT-Summ-OpenNMT}} A functional service based on the model is currently being integrated, as a summarization service, in the platforms Lynx \cite{morenoschneider2020j}, QURATOR \cite{rehm2020d} and European Language Grid \cite{rehm2020m}.


\section*{Acknowledgements}

The work presented in this paper has received funding from the European Union’s Horizon 2020 research and innovation programme under grant agreement no.~780602 (Lynx) and from the German Federal Ministry of Education and Research (BMBF) through the project QURATOR (Wachs\-tums\-kern no.~03WKDA1A).

\section{Bibliographical References}
\label{main:ref}

\bibliographystyle{./lrec}
\bibliography{./lrec2020}

\end{document}